\documentclass[10pt,twocolumn,letterpaper]{article}

\usepackage{cvpr}
\usepackage{times}
\usepackage{epsfig}
\usepackage{graphicx}
\usepackage{amsmath}
\usepackage{amssymb}
\usepackage{caption}

\graphicspath{ {figures/} }

\usepackage[breaklinks=true,bookmarks=false]{hyperref}

\cvprfinalcopy % *** Uncomment this line for the final submission

\def\cvprPaperID{****} % *** Enter the CVPR Paper ID here

\begin{document}

%%%%%%%%% TITLE
\title{A Study into the similarity in generator and discriminator in GAN architecture}

\author{Arjun Karuvally\\
University of Massachusetts, Amherst\\
{\tt\small akaruvally@cs.umass.edu}
%\and
%\\\tt\small secondauthor@i2.org}
}

\maketitle

\nocite{tensorflow2015-whitepaper}

%%%%%%%%% ABSTRACT
\begin{abstract}
   One popular generative model that has high quality results is the Generative Adversarial Networks(GAN). This type of architecture consists of two separate networks that play against each other. The generator creates an output from the input noise that is given to it. The discriminator has the task of determining if the input to it is real or fake. This takes place constantly eventually leads to the generator modelling the target distribution. This paper includes a study into the actual weights learned by the network and a study into the similarity of the discriminator and generator networks. The paper also tries to leverage the similarity between these networks and and shows that indeed both the networks may have similar structure with experimental evidence with a novel shared architecture.
\end{abstract}

%%%%%%%%% BODY TEXT
\section{Introduction}
	One of the popular generative Network architecuture is the Generative Adversaial Networks(GAN)\cite{2014arXiv1406.2661G}. It has shown promise in modelling well some target distributions. There are some really interesting and useful applications for these types of generative models. When the training dataset is small, these models can get around this issue by generating similar data using generative networks. Since its inception, training of GAN networks is known to be very difficult with a lot of issues that arises due to the decoupled training of generator and discriminator. One of these issues is called mode collapse, here the generator tricks the discriminator by mapping all the input noise into a single true output. This is undesirable as we require that the generator output to be diverse. The other issue that is encountered is when the discriminator network learns more than the generator and learns to exploit the weakness of generator by classifying all its output as false. This results in very less gradient propogation in the network and the generator fails to learn the target distribution. There has been networks that has achieved success in using these type of generative models to model target distributions. One of then is the DCGAN(Deep Convolutional GAN) networks\cite{DBLP:journals/corr/RadfordMC15}. Initially these type of networks were restricted to only some type of network architectures. Recently some studies has come up using a different loss function to achieve the same result and the paper argued that the different loss function results in a better GAN network and can incorporate different architectures. This is WGAN(Wasserstein GAN)\cite{2017arXiv170107875A}, but the main issue with this again is that it has a high training time and the results also does not have high quality. \\
	In this paper we delve into the correlation between the generator and discriminator and study if it is possible to leverage this property of the network to improve the training of GANs.

\section{Problem Statement}
	This paper discusses the similarity between the learned representations of the generator and discriminator networks. This is interesting because the possibility that we can leverage the correlations between networks can have consequences in improving the training time of the networks and also reduces the number of parameters to be used in network. This paper is based on the following hypothesis. The generator is trying to output a target distribution and  the discriminator is trying to determine if the input to it is from the target distribution or is generated from the generator. This means that the generator and the discriminator has to learn some features of the ditribution and the generator tries to model this features and discriminator tries to recognise these features.This has the implication that the discriminator learns some features corresponding to the higher level representations of the target distribution and the generator alse learns the same features. The paper delves into the possibility of using this similarity and proposes an architectures that leverages this property of GANs. Additionally this paper also intends to provide some qualitative and quantitative comparison of existing GAN architectures and the proposed method.
	
\section{Technical Approach}
	Initially to study the similarity in the learned weights of the network, I propose to take the learned weights from a fully trained DCGAN network and studying the similarity between the learned filters. Since two networks could learn filters at different positions, I take one filter and compare with all the learned filters. A similar experiment was used in CRelu\cite{DBLP:journals/corr/ShangSAL16} to show the existence of negatively correlated filters. The next portion is the study on creating a network architecture that leverages this similarity. For this I propose to initially test an architecture that has a shared middle layer. The hypothesis for using this type of architecture is that hopefully the common learned weights get accumulated in shared layers of the network. During backpropagation only the discriminator gets to update the weights. Experimentally it has proven that this type of architecture is better than allowing both the generator and discriminator train the weights.

\section{Similarity in Learned Weights}
	A DCGAN network was taken and a study is conducted on the similarity between filter weights in the generator and discriminator. The similarity metric is defined as the cosine similarity $max<\mu_t, \mu_{ref}>$ where $\mu_t$ is the target filter(in generator) for which we are finding similarity and $\mu_{ref}$ is the rest of the filters in the disciminator network. This is compared to the cosine similarity distribution that is produced by random weights. The results obtained are shown below.

\begin{figure}[h]
\begin{center}
\fbox{\includegraphics[width=0.8\linewidth]{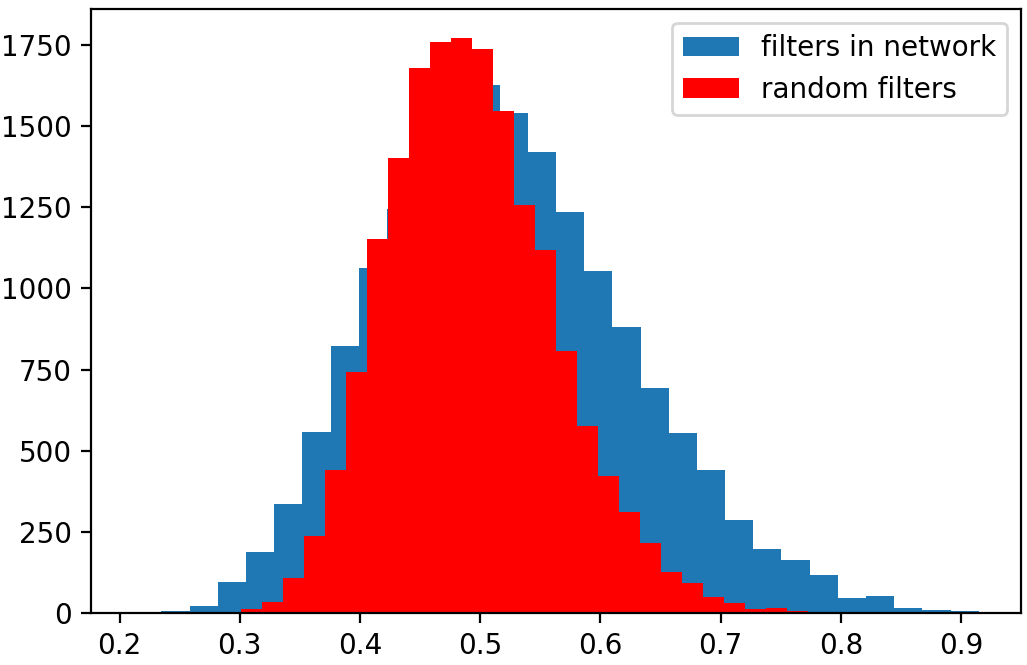}}
%\rule{0pt}{2in} \rule{0.9\linewidth}{0pt}
\end{center}
   \caption{Similarity distribution of weights from random filters and filters in the GAN networks. second last layer of generator to layer first layer of discriminator.}
\label{fig:long}
\label{fig:onecol}
\label{fig:similarity_1}
\end{figure}

The Figure ~\ref{fig:similarity_1} shows similarity between filters from the second last layer in generator to the first layer in discriminator. This shows that they are not correlated. What is interesting is the next correlation distribution.

\begin{figure}[h]
\begin{center}
\fbox{\includegraphics[width=0.8\linewidth]{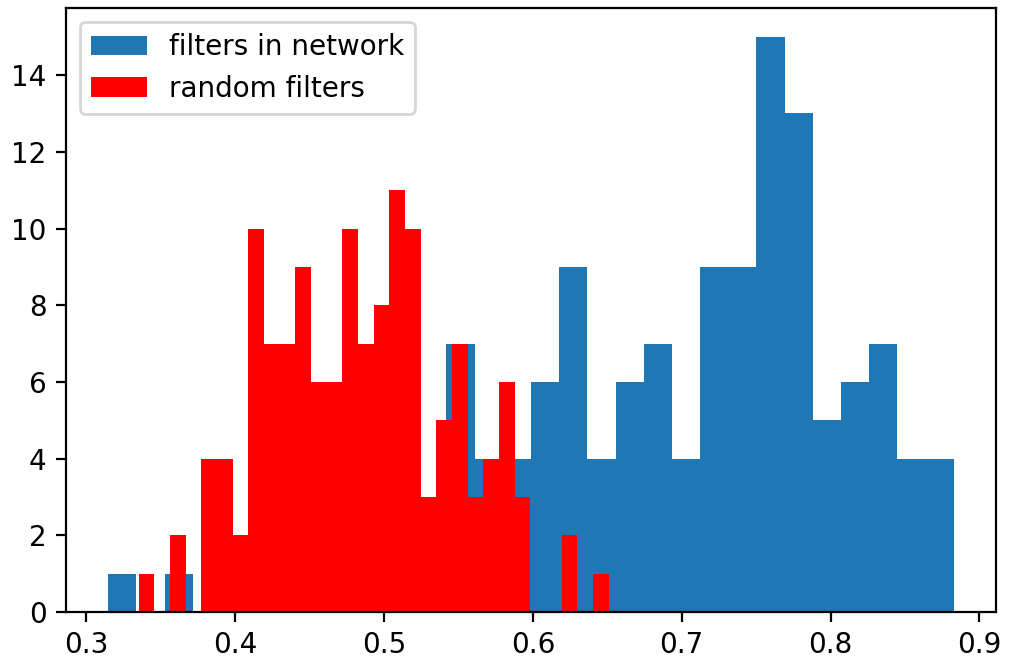}}
%\rule{0pt}{2in} \rule{0.9\linewidth}{0pt}
\end{center}
   \caption{Similarity distribution of weights from random filters and filters in the GAN networks. last layer of generator to the first layer in discriminator.}
\label{fig:long}
\label{fig:onecol}
\label{fig:similarity_2}
\end{figure}

In Figure ~\ref{fig:similarity_2}, it is interesting to see that the final layers in the generator are highly correlated with the first layer of the discriminator. Both these layers are the image facing side, meaning that generator creates image using this set of filters as last and the discriminator processes the image first using these filters. This test alone may not be enough to prove the claim that the structure may be the same. An architecture is proposed where the generator shares one layer of the disciminator to test this claim of similarity further.
\section{Architecture test on synthetic data}
\begin{figure}[h]
\begin{center}
\fbox{\includegraphics[width=0.8\linewidth]{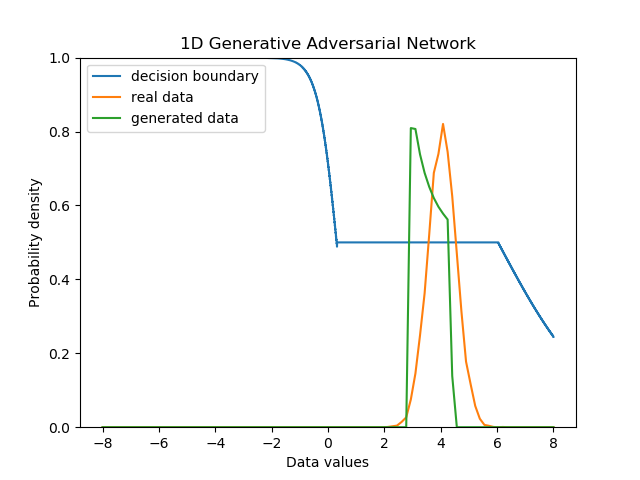}}
%\rule{0pt}{2in} \rule{0.9\linewidth}{0pt}
\end{center}
   \caption{The output produced by the proposed GAN implementation(shared weights). The output captured the mean and standard deviation well in this. Notice that the behavious of the disciminator is very different.}
\label{fig:long}
\label{fig:onecol}
\label{fig:synthetic_data_shared}
\end{figure}

	The proposed architecture was taken and applied on a synthetic dataset that is used widely in GANs. The objective in this is to model a gaussian distribution when the input noise given to generator is a uniform distribution. The baseline result has a higher quality but the proposed framework converged very quickly to the desired distribution. Analysing the loss function, it was found that using the shared weights from the discriminator, the generator was able to capture the mean and variance quickly then it proceeded to small changes in output distribution. It was also noticed that the generator quickly captured the range of the target distribution such that the discriminator was not given much data points beyond the target distribution to learn if it is fake or not. This leads to the real value output beyond the range of the target distribution in Figure ~\ref{fig:synthetic_data_shared}.

\begin{figure}[h]
\begin{center}
\fbox{\includegraphics[width=0.8\linewidth]{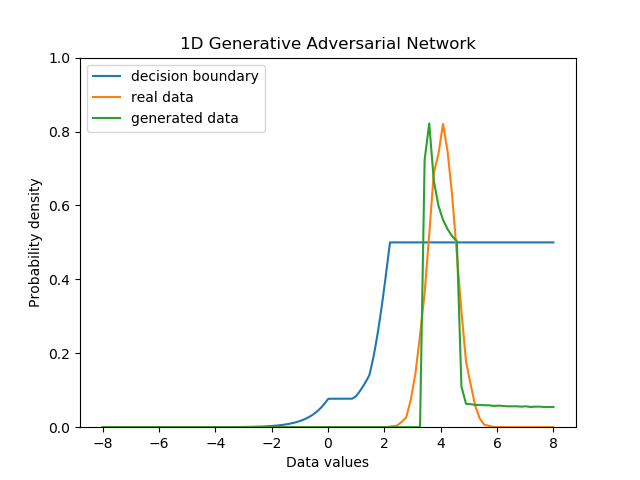}}
%\rule{0pt}{2in} \rule{0.9\linewidth}{0pt}
\end{center}
   \caption{The output produced by the tuned native GAN implementation. Notice the long outlier region outside the target distribution.}
\label{fig:long}
\label{fig:onecol}
\label{fig:synthetic_data_baseline}
\end{figure}

These experiments show that the architecture may be feasible to be deployed on bigger problems. Next experiments focus on testing the architecture on MNIST and CelebA datasets.

\section{Experimental Results}
There are some difficulties that are encountered when we try to create a new architecture using the GAN architecture.\\
It is known that training GAN networks is highly unstable. Hyperparameter optimization is somewhat difficult for GAN networks as we have to find the right balance between learning rates to efficiently train the network. Due to these issues the WGAN objective is used for the implementation. WGAN is proven to be more stable on varying architectures. The learning rates were set at $ 5 \times e^{-5} $. Two models are trained. One using WGAN objective without using the shared architecture. This is considered as the baseline. The same architecture is trained while sharing the weights of the second convolution layer. The second layer weights were chosen by the intuition that both the generator and discriminator maps to the same feature dimension hence the mapping to the next one could be same and the generator could reuse the same transformations from discriminator. Experimentally, it is seen that using some of the initial layers leads to better results compared to sharing some final layers also most intial layers produce outputs with similar quality. Sharing multiple layers lead to drastic drop in training quality and there were issues with balancing the training of generator and discriminator.

\subsection{MNIST}
The figures show the results that were obtained on the MNIST dataset.

\begin{figure}[t]
\begin{center}
\fbox{\includegraphics[width=0.8\linewidth]{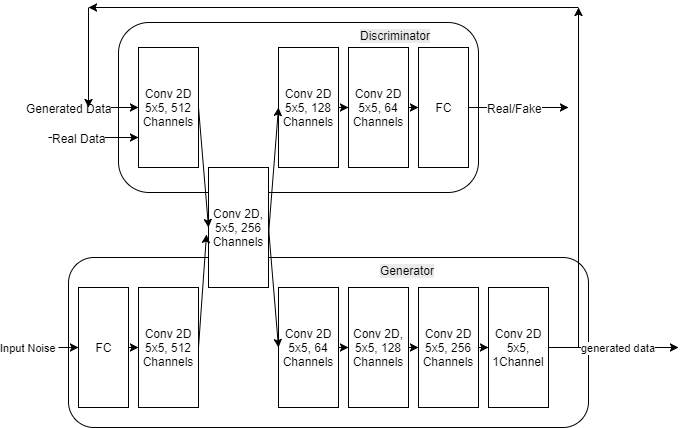}}
%\rule{0pt}{2in} \rule{0.9\linewidth}{0pt}
\end{center}
   \caption{The shared layer architecture.}
\label{fig:long}
\label{fig:onecol}
\end{figure}

\begin{figure}[h]
\begin{center}
\fbox{\includegraphics[width=0.8\linewidth]{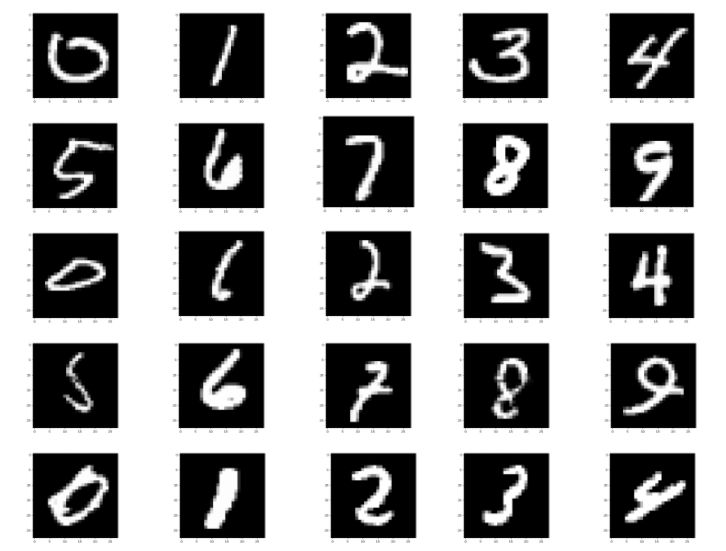}}
%\rule{0pt}{2in} \rule{0.9\linewidth}{0pt}
\end{center}
   \caption{Some samples drawn from the dataset}
\label{fig:long}
\label{fig:onecol} 
\end{figure}

\begin{figure}[h]
\begin{center}
\fbox{\includegraphics[width=0.8\linewidth]{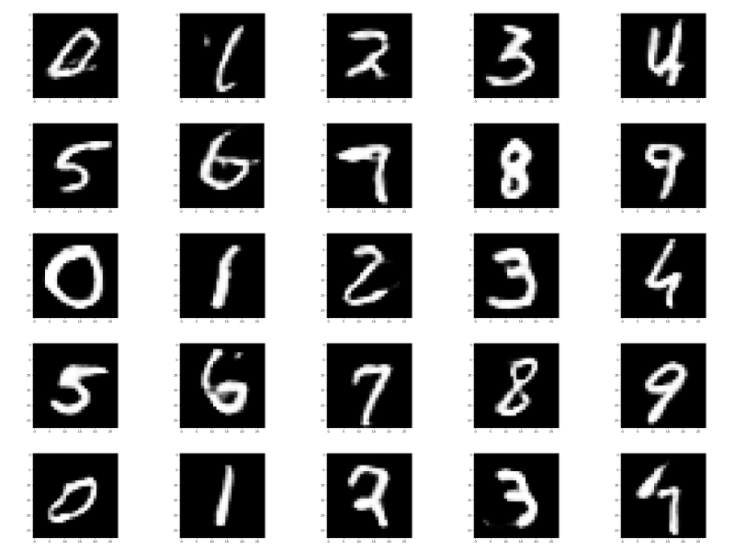}}
%\rule{0pt}{2in} \rule{0.9\linewidth}{0pt}
\end{center}
   \caption{Some samples drawn from the output of baseline architecture after 10th epoch.}
\label{fig:long}
\label{fig:onecol}
\end{figure}

%\newpage

\begin{figure}[h]
\begin{center}
\fbox{\includegraphics[width=0.8\linewidth]{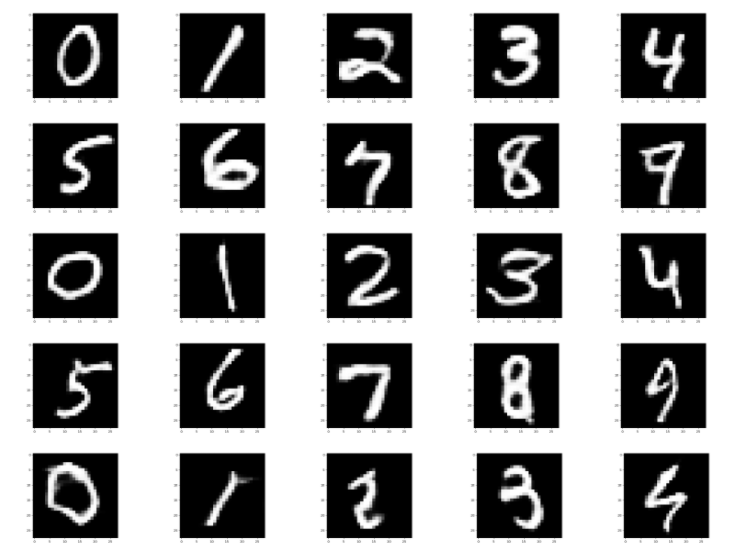}}
%\rule{0pt}{2in} \rule{0.9\linewidth}{0pt}
\end{center}
   \caption{Some samples drawn from output of shared architecture after 10th epoch.}
\label{fig:long}
\label{fig:onecol}
\end{figure}

It can be noted that the shared model was able to capture the distribution well just like the baseline architecture. The difference in terms of quality of both shared and baseline are less, and both were trained till the same number of steps. Due to the lack of proper quantitative measure of the quality of generated models, it is quite difficult to tell if one model performs well compared to another one.

\subsubsection{Evaluation}
GAN models lack an objective function, so it is diffucult to compare performance of different models. One intuitive performance metric is to use human evaluators to evaluate the quality of sampled. As it is shown in paper by Tim Salimans et.al.\cite{DBLP:journals/corr/SalimansGZCRC16} in usage of human evaluators, the metric varies with respect to the motivation and the setup of the task, and a high number of human evaluations may be required to correctly ascertain the performance of a model. \\
To get over this roadblock the paper suggests an alternative. The inception model(high accuracy model) is taken to get the conditional label distribution $p(y|x)$. Images with meaningful distribution has low entropy for the conditional distribution. They add another metric to determine if the generative model produces varying outputs(no mode collapse). I believe that adding this does not give a quantitative measure of just the quality of generated images. I believe that both measures should be treated different. So to determine the quality of samples from the MNIST dataset, a high accuracy classification model was trained. The test accuracy for the classification task obtained for the model was 99.23\%. This is used to find the distribution of entropy values for 1000 samples drawn from dataset, baseline and shared model. In addition to this the entropy distribution for the initial model(without training) is also added to show how the entropy distribution of the learned models changes.

\begin{figure}[h]
\begin{center}
\fbox{\includegraphics[width=0.8\linewidth]{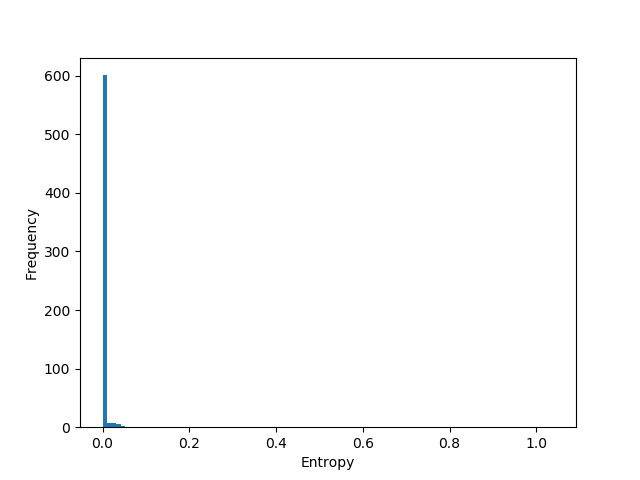}}
%\rule{0pt}{2in} \rule{0.9\linewidth}{0pt}
\end{center}
   \caption{The distribution of entropy values for samples from dataset.}
\label{fig:long}
\label{fig:onecol}
\label{fig:mnist_dataset_entropy}
\end{figure}

\begin{figure}[h]
\begin{center}
\fbox{\includegraphics[width=0.8\linewidth]{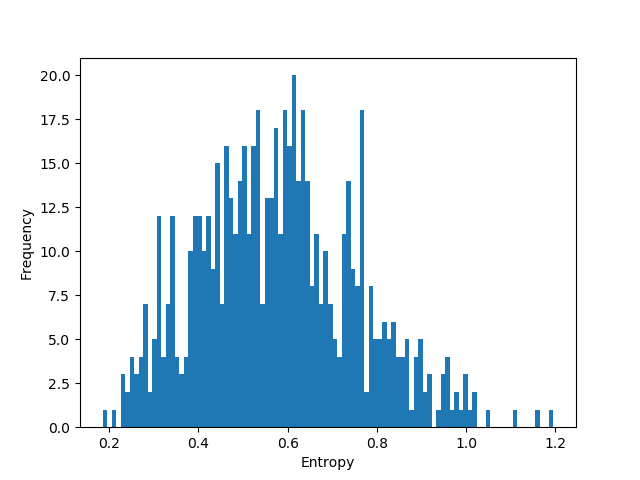}}
%\rule{0pt}{2in} \rule{0.9\linewidth}{0pt}
\end{center}
   \caption{The distribution of entropy values for samples from initial(untrained) model.}
\label{fig:long}
\label{fig:onecol}
\label{fig:mnist_dataset_random}
\end{figure}

\begin{figure}[h]
\begin{center}
\fbox{\includegraphics[width=0.8\linewidth]{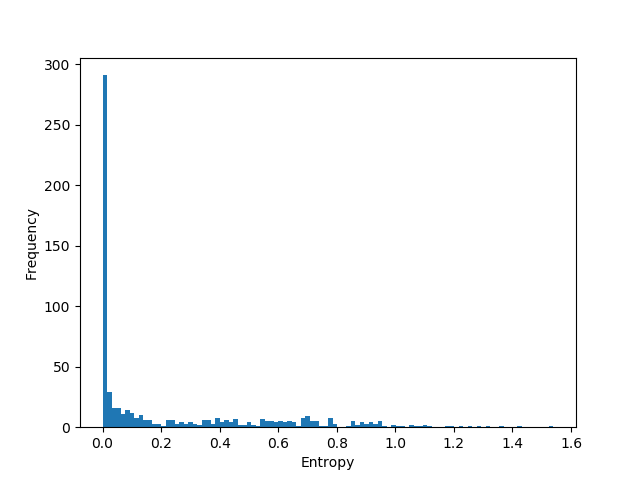}}
%\rule{0pt}{2in} \rule{0.9\linewidth}{0pt}
\end{center}
   \caption{The distribution of entropy values for samples from baseline WGAN.}
\label{fig:long}
\label{fig:onecol}
\label{fig:mnist_dataset_baseline}
\end{figure}

\begin{figure}[h]
\begin{center}
\fbox{\includegraphics[width=0.8\linewidth]{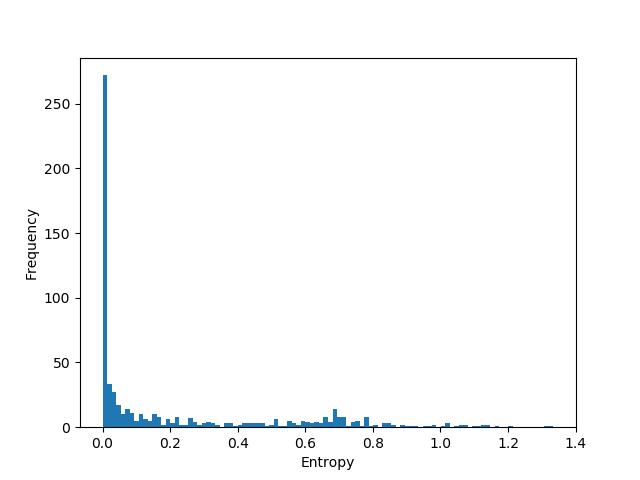}}
%\rule{0pt}{2in} \rule{0.9\linewidth}{0pt}
\end{center}
   \caption{The distribution of entropy values for samples from shared WGAN model.}
\label{fig:long}
\label{fig:onecol}
\label{fig:mnist_dataset_shared}
\end{figure}

It can be seen from  the Figure ~\ref{fig:mnist_dataset_entropy} that the entropy distribution for samples from the dataset is concentrated near zero. This implies that most images have near to zero entropy label distribution. The entropy of initial generated images as shown in Figure ~\ref{fig:mnist_dataset_random} has high entropy with high mean and variance. 
After learning the baseline(Figure ~\ref{fig:mnist_dataset_baseline}) and shared(Figure ~\ref{fig:mnist_dataset_shared}) models changes the entropy distribution from a noisy high mean distribution to means closer to zero. The generative models are not perfect, so not all values are very close to zero.

Below the mean, standard deviation metrics of the models are shown.
\begin{itemize}
	\item Dataset
		\begin{itemize}
			\item Mean Entropy:  0.00621325
			\item Standard Deviation:  0.0478745
		\end{itemize}
	\item Initial model
		\begin{itemize}
			\item Mean Entropy:  0.481903
			\item Standard Deviation:  0.160115
		\end{itemize}
	\item Baseline WGAN model
		\begin{itemize}
			\item Mean Entropy:  0.208024
			\item Standard Deviation:  0.310353
		\end{itemize}
	\item Shared WGAN model
		\begin{itemize}
			\item Mean Entropy:  0.209418
			\item Standard Deviation:  0.30961
		\end{itemize}
\end{itemize}

It is to be noted here that the values shown are not absolute measure of quality. Using these metrics for loss function has proven to be unsuccessful, but it is proven that the entropy based evaluation corresponds well with the quality of human evaluations.

\subsection{CelebA}
\begin{figure}[h]
\begin{center}
\fbox{\includegraphics[width=0.8\linewidth]{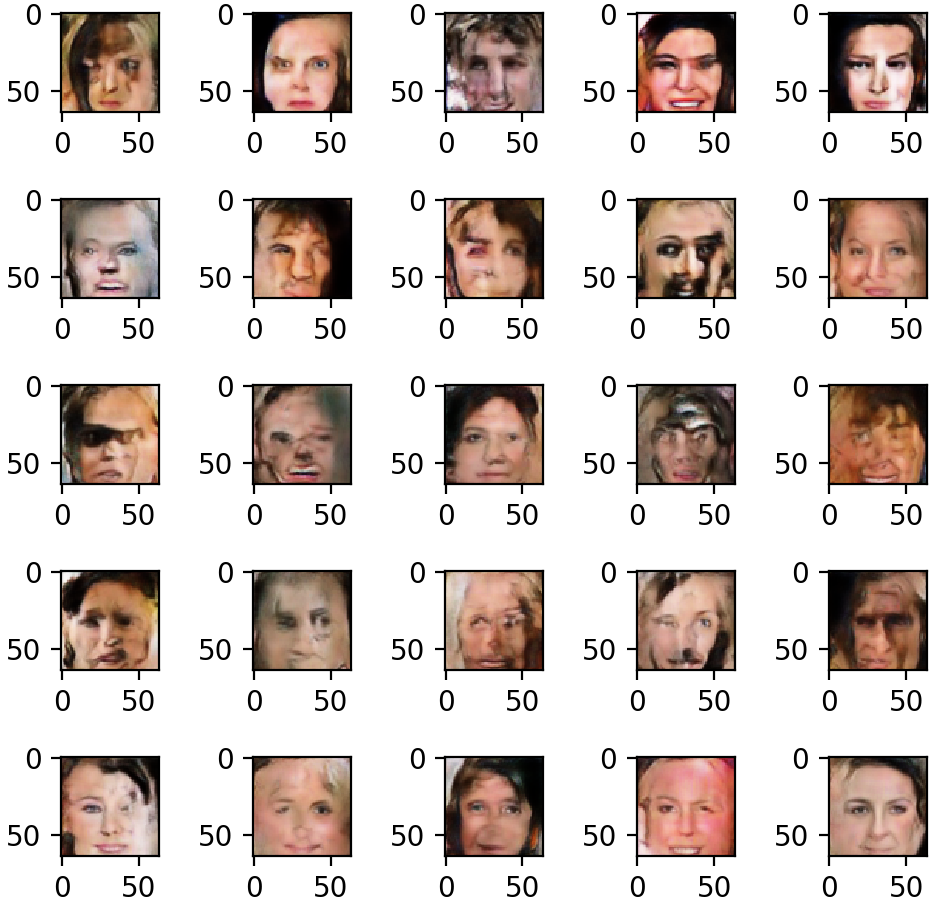}}
%\rule{0pt}{2in} \rule{0.9\linewidth}{0pt}
\end{center}
   \caption{Samples from shared WGAN model trained on the CelebA dataset.}
\label{fig:long}
\label{fig:onecol}
\label{fig:celebA_samples}
\end{figure}

The shared architecture is takan and modified to accomodate the change in number of channels without any change in the capacity of the model. This is tested on the celebA dataset. It can be seen in Figure ~\ref{fig:celebA_samples} that the quality of the generated faces are very less but the generator was able to reproduce the general structure of faces evident from this output. The present best generative model, the DCGAN network could produce images with high quality.

\section{Conclusion}
It can be seen that although the proposed architecture was not able to achieve state of the art results of GAN architectures, it was able to achieve results with quality similar to the baseline WGAN with the similar architecture. Creating a shared layer, it is expected that the generator would perform quite poor compared to the disciminator during training and the training would not proceed well. This is especially true in case of GAN networks as the sharing produces a moving target for the generator that it cannot bypass like ordinary networks that if we introduce noise in one of the layers and learn the other weights, it learns to bypass the noise. It could mean that the tasks has some underlying similarity that could be exploited in future and with more testing on different architectures could result in surpassing the current state of the art in GAN architectures.

\section{Future Work}
In future I propose more experimentation on this shared architecture. Including other features to stabilize networks like feature matching, minibatch discimination, historical averaging, virtual batch normalization that have been proven to improve the quality of outputs produced by GAN networks. It is also shown from crude experimentation that the proposed modification is stable on changing architecture as long as the shared weights have same size and shape. This effect is worth looking into and more experimentation with baseline is suggested to evaluate this claim. This suggested architecture does not have proper theoretical support. In future work this experimental results have to be backed proper theoretical 
evidence. Another interesting area to look into is that in GAN networks usually disciminator is discarded after training of GAN network but I believe that the disciminator has some rich features learned by telling both distributions apart that could be used for initializing the training of a regression or classification model on a dataset and analyse how we could improve training.

\footnote {Source code for experiments: \href{https://github.com/arjun23496/Shared-WGAN}{https://github.com/arjun23496/Shared-WGAN}} 
%\begin{figure}[t]
%	\includegraphics[scale=0.75]{architecture.png}
%\end{figure}
{\small
\bibliographystyle{ieee}
\bibliography{egbib}
}
\end{document}